# Using Evolutionary Design to Interactively Sketch Car Silhouettes and Stimulate Designer's Creativity

François Cluzel[a], Bernard Yannou[a,1], Markus Dihlmann[b]

[a]Laboratoire Genie Industriel, Ecole Centrale Paris
Grande Voie des Vignes, 92290 Chatenay-Malabry, France
E-mail: {francois.cluzel, bernard.yannou}@ecp.fr
Phone: +33 1 41 13 15 21

[b]Institute of Applied Analysis and Numerical Simulation, University of Stuttgart
Pfaffenwaldring 57, 70569 Stuttgart, Germany
E-mail: dihlmann@mathematik.uni-stuttgart.de

***Abstract***: An Interactive Genetic Algorithm is proposed to progressively sketch the desired side-view of a car profile. It adopts a Fourier decomposition of a 2D profile as the genotype, and proposes a cross-over mechanism. In addition, a formula function of two genes' discrepancies is fitted to the perceived dissimilarity between two car profiles. This similarity index is intensively used, throughout a series of user tests, to highlight the added value of the IGA compared to a systematic car shape exploration, to prove its ability to create superior satisfactory designs and to stimulate designer's creativity. These tests have involved six designers with a design goal defined by a semantic attribute. The results reveal that if "friendly" is diversely interpreted in terms of car shapes, "sportive" denotes a very conventional representation which may be a limitation for shape renewal.

***Keywords***: interactive genetic algorithm, evolutionary design, shape design, subjective evaluation, user tests, car profile.

## 1. Introduction

The art of sketching, and the sketches themselves, are considered to be an important part of conceptual design as well as the way for designers to condense their knowledge and exploration [1-3]. In their work, Yang et al. [4,5] address the correlation between the quantity of generated sketches and their impact on design outcomes. Moreover, they explore the relationship between the moment of sketching in the design process, especially early in the design, and the impact on the outcomes. Based on the correlation coefficient exploration, statistically it is found that the quantity of concept generation is significantly correlated with project outcome quality. This is also the case for the timing of sketches: Yang [3] found that early sketching and prototyping gave the best project outcome qualities. Moreover, some studies focus on the importance of the prototyping stage. Yang [6] explores the influence of the prototype complexity, prototype quantity and the time spent on this activity and their relationship with the design outcome. The results of the study show that the lesser number of parts in a device, then the greater the project outcome quality.

Evolutionary Computation (EC) has become a major approach for the exploration and evaluation of design solutions, and especially 2D sketched solutions. An EC method basically uses the genetic algorithms (GA) [7,8], which were originally used to find solutions for complex optimization problems. For example, Poirson et al use GAs to optimize the design of brass musical instruments by considering mathematical and perceptual objectives [9]. Taking the evolution in nature as a paradigm, the GAs work on the basis of a population of individuals, where each individual represents a possible solution for the initial problem. The structure and qualities of each individual are encoded in their genomes. Through the recombination of these genomes individuals can reproduce themselves and produce new individuals (solutions), and through a sort of natural selection, individuals who are not adapted to the environment (whose properties do not satisfy the expectations) are not selected for procreation. In this way, individuals display better and better qualities over the generations. Interactive Genetic Algorithms (IGA, see [10,11]) represent a special class of GAs where a human (here, the style designer) is a key player embedded within the task of selecting individuals from a generation. IGAs are then particularly adapted to situations where it is impossible to explicitly express a preference function (the fitness function) on individuals or even when it is hard to qualify expected properties. This is typically the case with style designers.

---
[1] Corrresponding author

A major difficulty when using GAs in automatic design systems is the encoding of the genome (see [10]), which means the way of coding the individual's phenotype (physical structure) into the genotype (genome). Most systems use a direct encoding where the geometrical dimensions and structures of the design object are directly represented in the genome (see for example [11]). For example, when designing a bottle [12,13] or when finding a design for cylindrical shapes [14], the phenotype is represented in the genome by a sequence of geometrical parameters such as the radii, lengths and part locations. Consequently, encoding is context dependent. Other works use tree structures [15] or shape grammars [16] to encode the genome. Kim and Cho [17] have used a set of predefined parts of clothes to find new fashion designs by recombining these parts. In addition, all of these systems are conceived for a given design domain. Implementing these methods in new design fields is a difficult and time consuming process. However, a good design method should be applicable, as much as possible, to a large spectrum of situations.

In this paper, we first propose an encoding method for a 2D-closed-curve which is supposed to meet a desired style. This method can be applied to all possible objects represented by their 2D-silhouettes. For instance, a car silhouette or profile is an essential style feature for a car. Indeed, Cheutet [18] has shown that the character of a car profile is primarily expressed through a series of about ten lines (see Figure 1). Five of them: hood line, windshield line, roof line, wheelbase line and wheel arch, may be merged into a silhouette closed line. It has been proved that these lines, and especially the silhouette, have a strong determining influence on the perception of the car while embedding perceptual attributes such as: sportiveness, aggressiveness or peacefulness, etc.

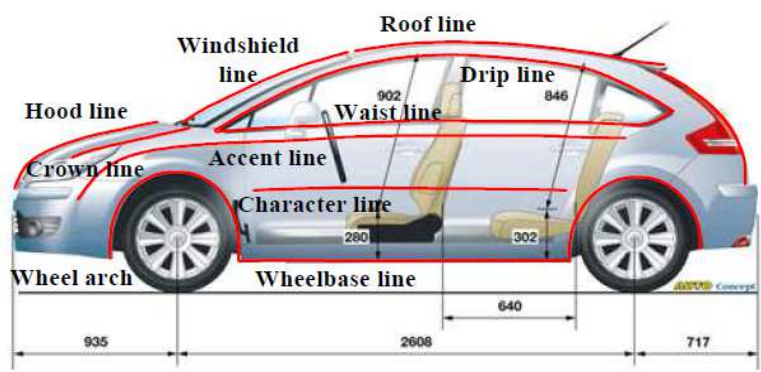

**Figure 1. Cheutet models a car using key lines [18]**

In addition, it has been proved that the aesthetic aspects of a car amounts for 70% of customers purchase intents [18]. Other approaches consider car shape design in a tool for assisting the designer. Petiot and Dagher [19] propose a tool to evaluate car front-end designs through semantic attributes (see Figure 2). Osborn et al. [16] use shape grammars to assist the user in the design of new car profiles (see Figure 3). Kelly et al. [11] describe a car silhouette with 12 points (8 fixed points and 4 varying points) and use an IGA to find new designs (see Figure 4).

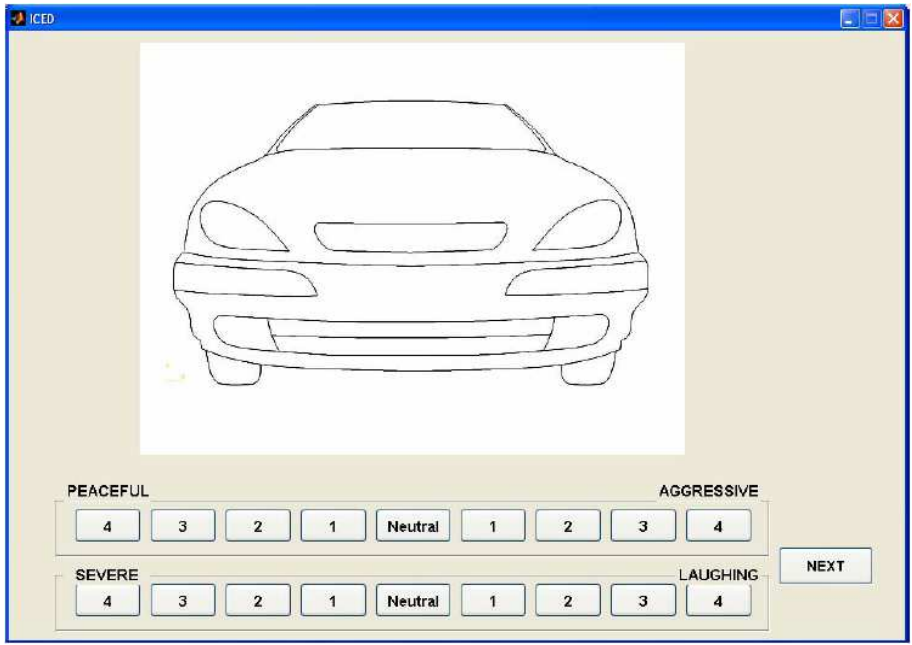

**Figure 2. Perceptual evaluation of car front-end designs by Petiot and Dagher [19]**

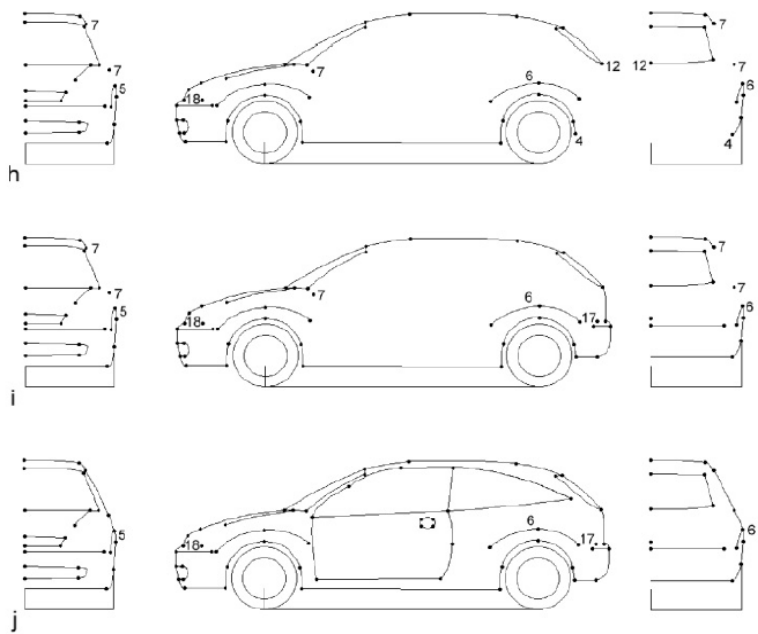

**Figure 3. Some stages to build a Ford Focus using the shape grammars defined by Osborn et al. [16]**

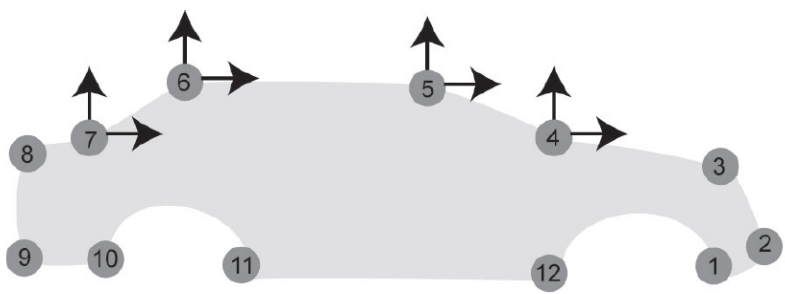

**Figure 4. Car silhouette modeling from Kelly et al. [11]. Location of points 4, 5, 6 and 7 vary in the vertical and horizontal plane. The other points are fixed.**

Consequently, it turns out that the design of car shapes or silhouettes has been studied with different techniques, one of them - Kelly et al. [11] – is even with an IGA. In this paper we propose an IGA based on the encoding of the car silhouette genes after a Fourier decomposition. In section 2, we provide some arguments to compare this GA modeling technique with parameterized curves and shape grammar approaches. Next, the Interactive GA (IGA) is presented and the crossing-over operation between genes detailed. The interactivity consists in allowing a style designer to qualitatively assess individuals at each generation. As such, new innovative designs are expected to emerge through a balanced collaboration between an automatic design space exploration process and the interaction of a designer. In section 3, we propose a method to identify an index of a perceived similarity between two car profiles. In section 4, we use this index through a series of user tests to prove that innovation and surprise may emerge from this process in an effective way. Section 5 concludes on the properties of the IGA approach for innovative sketching.

## 2. Model description

### 2.1. Quality of an encoding model

In this work, we have decided to approximate the visual perception of a car through a 2D "side-view" closed curve. Mathematically, we define $\mathcal{M}$, the set of closed curves in $\mathbb{R}^2$. We define the quotient set $\mathcal{M}\backslash\mathcal{R}$ where $\mathcal{R}$ is the following equivalence relation: "Two curves of $\mathcal{M}$ are equivalent with respect to $\mathcal{R}$, if and only if, they are similar through rotations, homotheties and translations".

Numerically, an equivalence class can be described in different ways (resampled curve, Fourier spectrum, wavelet coefficients…). The book of Mallat [20] discusses these different modeling techniques thoroughly. These are different surrogate representations for a same object; each of which have their advantages and disadvantages.

Considering their ability to be appropriate to encode closed curves for an evolutionary design purpose, three properties must be characterized:

- Ability to avoid the emergence of "degenerated individuals" (aberrations are likely to be invisible for frequential representation) (improving "consistency"),
- The ability to define a convenient cross-over mechanism,
- The ability to define a convenient mutation mechanism.

Finding a way to efficiently represent car silhouettes is an essential point in our study. This issue is closely related to the notion of ontology. Indeed, we need a formal representation of our knowledge in order to decide how to manipulate our objects while remaining in the restricted field of car design. Thus, we must ensure that, for example, crossing two vehicles will create a new object that will still be an "acceptable vehicle", i.e. an apparently possible vehicle. The main principles to design an efficient ontology have already been investigated by researchers such as Tom Gruber [21]. Gruber gives some important dimensions that one must take into consideration when designing a new ontology:

- *Clarity:* In our case, this implies that our representation must be objective and is not context dependent.
- *Consistency:* A very consistent ontology should not be able to create "monsters" from valid rules. However in our case, too much consistency will constrain creativity.
- *Extensibility:* One should keep in mind that extensions are likely to be added in the future.
- *Weak coding distortion:* When we are representing a real object, such as a car, there is always some "distortion" due to the fact that our ontology cannot represent all sides of the object. An essential drawback of 2D descriptive methods (cf. infra) is their strong distortion.
- *Ontological complexity:* Our representation must be parsimonious.

It must be kept in mind that the main goal of ontologies is, above all, to conceptualize a "fixed" field of knowledge. A trade-off between the ontology "quality" and the potential creativity is essential in our work.

Table 1 gives some example of such ontologies.

|  | Completeness | Robustness | Consistency | Extensibility | Adaptivity to genetic algorithms |
|---|---|---|---|---|---|
| **Shape grammars** | ++ | ++ | ++ | - | -- |
| **Parameterized contour** | -- | -- | -- | ++ | ++ |
| **Fourier harmonics** | -- | ++ | -- | ++ | ++ |
| **Wavelets** | -- | ++ | - | ++ | ++ |

**Table 1. Comparison of different encoding methods**

Many ontologies are based on taxonomic relationships: entities, such as cars, are classified in different families organized in a global hierarchical structure. A car can thus be described as a special realization of some taxonomic scheme. This method leads to generative models such as shape grammars [16]. Likewise, an approach based on Fourier harmonics is not as rigid as the "classical" approach by contour parameterization for multiple reasons:

- This encoding is supposed to embrace a much vaster space of possible 2D-closed-curves – or 2D-silhouettes – than by a parameterization approach whose risk is not to be able to propose "out of the box" solutions;
- All kind of shapes may be represented, even with small details that can be of the highest importance for provoking feelings and emotions;
- Encoding may be performed through a constant length of the genotype, which greatly simplifies crucial GA stages such as the cross-over operation between parent individuals;
- Finally, the genes in our solution have proved to be narrowly associated with apparent characteristics that must converge after several generations to the ideal 2D shapes.

For these reasons we decided to use a simplistic ontology based on Fourier harmonics, which is described in the next part and which is known as the "purely descriptive model".

### 2.2. Fourier harmonics to encode car silhouettes

McGarva [22] has proposed a method for coding the phenotype of 2D-closed-curves using their development into a Fourier series. Personally, we have already used this theory in [23] for encoding a 2D-closed-curve into the five first Fourier harmonics of this decomposition.

The McGarva's theory of Fourier decomposition of a closed curve [22] considers that the position of each point belonging to this curve can be expressed by a complex function in the complex plane, as written in formula (1):

$$z(t) = x(t) + i\, y(t) \tag{1}$$

As the curve is closed, z(t) is a periodic function. The period is normalized with: *z(t+1) = z(t)*. *z(t)* can be developed into a Fourier series in formula (2).

$$z(t) = \sum_{m=-\infty}^{\infty} a_m \exp(2\pi i m t) \tag{2}$$

where the complex Fourier coefficients can be calculated by formula (3).

$$a_m = \int_0^1 z(t) \exp(-2\pi i m t)\, dt \tag{3}$$

Coefficient $a_0$ is called the fundamental, $a_1$ and $a_{-1}$ represent the first harmonic, $a_2$ and $a_{-2}$ the second harmonic, etc.

As we will see later, the function $z(t)$ is not known as an explicit function from the beginning. Instead, we assume that the curve has been initially defined by a set of successive points $z_k$ ($k=0,..,N$) which belong to the curve. So, in order to calculate the $a_m$ coefficients (3) a numeric approximation is required. This approximation is obtained by dividing the curve into $N$ segments connecting each point with its successor. We call $t_k$ the length of the curve between the first point $z_0$ and the point $z_k$. Under these conditions the integral can be calculated by the trapezium formula (4):

$$a_m = \sum_{k=0}^{N} \left(\frac{t_{k+1}-t_k}{2}\right)(z_{k+1}\exp(-2\pi i m t_{k+1}) + z_k\exp(-2\pi i m t_k)) \tag{4}$$

where $z$ is a periodic function, ($z_{N+1} = z_0$).

The value of $t_k$ is the ratio of the length of the curve to point $k$ and the total length of the curve (see formula 5).

$$t_k = \frac{L_k}{L},\ L = \sum_{i=0}^{N}\sqrt{(x_i-x_{i+1})^2 + (y_i-y_{i+1})^2}\ \text{and}$$

$$L_k = \sum_{i=0}^{k-1}\sqrt{(x_i-x_{i+1})^2 + (y_i-y_{i+1})^2} \tag{5}$$

where the total length $L$ is the sum of the lengths of all segments, $L_k$ is the length from the origin to the current point, $x_{N+1} = x_0$ and $y_{N+1} = y_0$.

The construction of the genome of an object is made by developing its silhouette into a Fourier series and defining the fundamental (the coefficient $a_0$) as gene number zero. The first harmonic ($a_1$, $a_{-1}$) will be called the first gene, the second harmonic the second gene, etc.

On the basis of the genome, the original shape of the individual can be reconstructed. Each point $P_k$, with coordinates ($x_k$, $y_k$) on the curve $z^*$ which approximates the silhouette of the car, can be calculated by formula (6).

$$z^*(t_k) = x_k + i y_k = \sum_{m=-p}^{p} a_m \exp(2\pi i m t_k) \tag{6}$$

where $t_k$ ($0 \le t_k \le 1$) is the position on the curve and $p$ fixes the number of harmonics used for the decoding. The more harmonics used for the decoding the more precise the approximation of the original curve (as seen in Figure 5). $p$ is called the decoding "precision".

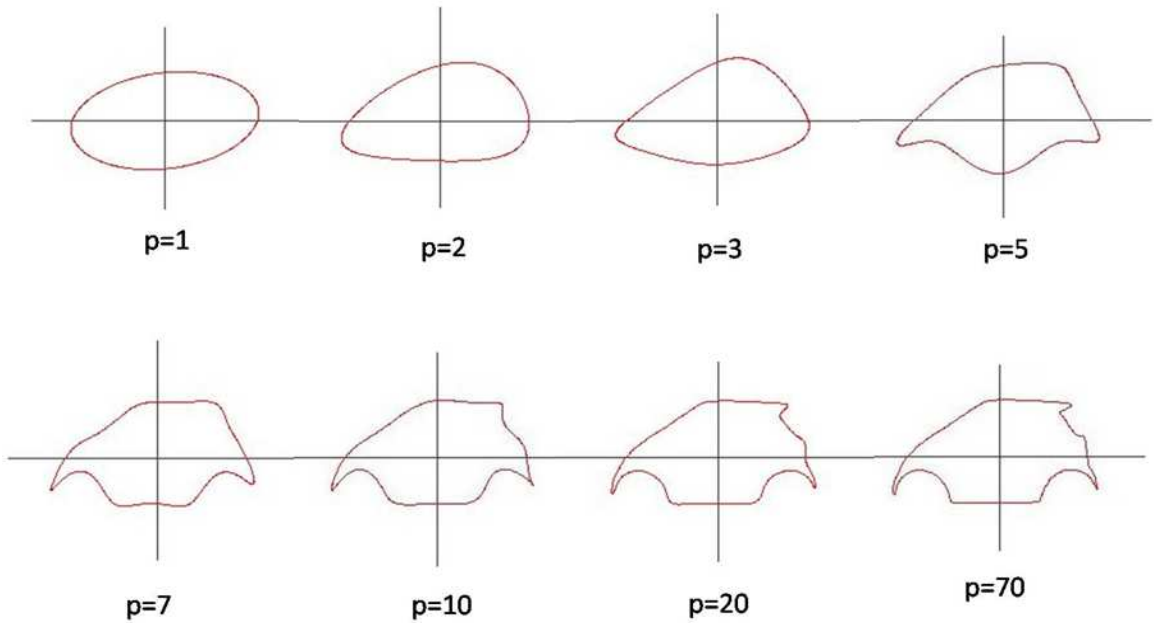

**Figure 5. Decoding a genome of a Smart car with different precisions**

The encoding of individuals allows an initial population that further evolves along several generations, depending on user choices, to be created. This process is explained in the next section.

### 2.3. An IGA-based model

The IGA model is divided into two phases, as shown in Figure 6. An initial population is created during Phase 1, while Phase 2 concerns the evolution of the population through multiple generations.

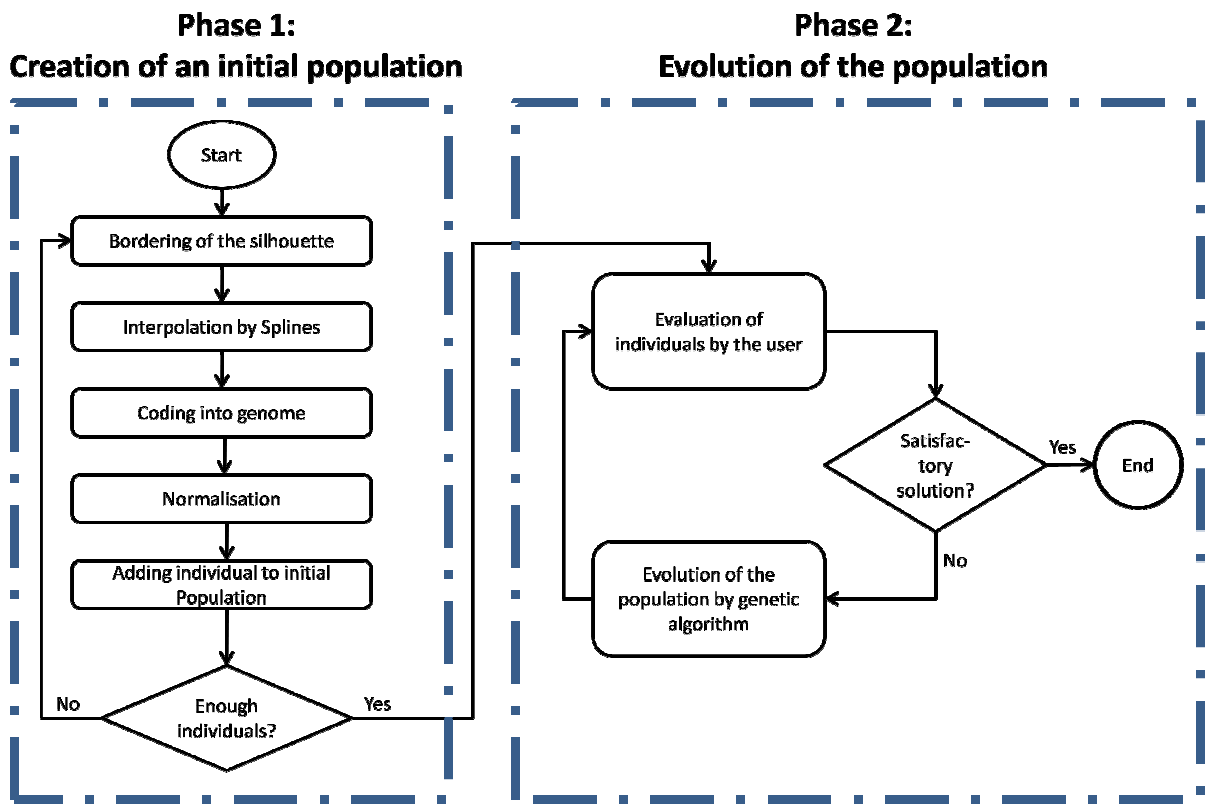

**Figure 6. Diagrammatic plan of the IGA process**

A genetic algorithm requires an initial population. In our case, this population consists of 30 car body silhouettes which already exist. These cars have been chosen to represent a large range of possible forms (different car brands, types or ages). To easily sketch this population, a Java interface has been programmed to draw curves on a plane and code them into a genome. From a picture of the car, the user has the possibility to draw a contour-chart around it by clicking on the screen. We obtain a closed curve representing the silhouette of the car (see Figure 7). In our experience a number M of 60 to 80 points are required to obtain satisfactory results, that means to visually capture the finest details of a car silhouette. However, as this "hand made" silhouette is a non smooth curve we would obtain a high numerical approximation error when applying the trapezium formula (4) to calculate the integral (3). This would lead to a very inaccurate representation of the silhouette and would result in high oscillations when trying to reconstruct the curve with (6) (see Table 2). That is why we increase the number of points by smooth interpolations, using bicubic splines linking three successive points (see Figure 8). Thereby we finally obtain a number of points $N$ describing the silhouette by a smooth curve where $N>M$.

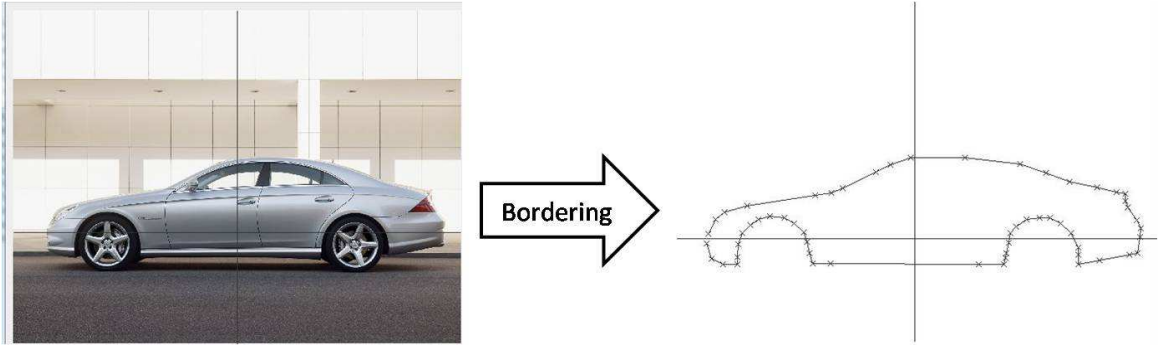

**Figure 7. After bordering we obtain a closed curve representing the car silhouette of an existing car**

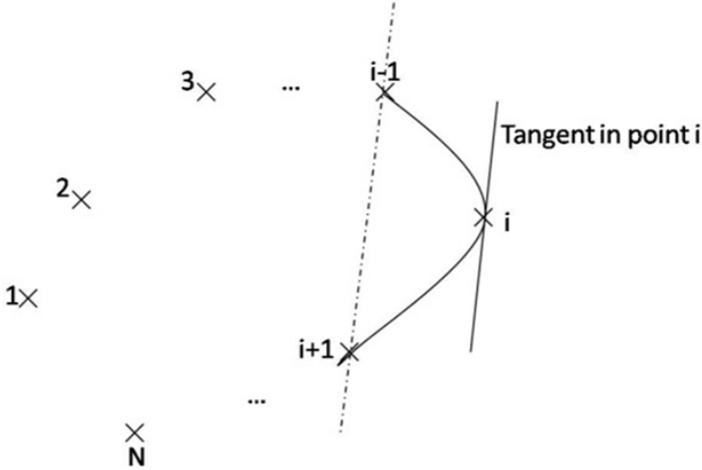

**Figure 8. The tangent of the spline at point i is parallel to the line passing through points i-1 and i+1**

A good encoding quality is reached if the phenotype obtained after the encoding-decoding process is visually similar to the initial profile. In our model, it corresponds to finding a satisfactory balance between the number $N$ of points on the curve used for coding and the number $p$ of the harmonics used when decoding the genome into a curve. Indeed, the higher the number of harmonics used for decoding the more precise the approximation to the original curve. On the other hand, the number $N$ of points on the curve used for coding the genome has an influence on the precision of the Fourier coefficients. A simple full factorial design of experiments (see Table 2) has been performed to identify the best compromise between $N$ and $p$. The evaluation of the fidelity level of the complete encoding-decoding process has been qualitatively – visually - performed by us on a 5-level scale {I – inaccurate, O – oscillations, SO – strong oscillations, G – Good result, GG – very good result}. In definitive, we found that a satisfactory choice was achieved with a genome size of 71 and a number *N* of approximately 1500 points for the interpolation since both initial and resulting silhouettes were visually identical.

| p\N | 80 | 100 | 200 | 500 | 700 | 1000 | 1200 | 1500 | 2000 |
|---|---|---|---|---|---|---|---|---|---|
| 5 | I | I | I | I | I | I | I | I | I |
| 7 | I | I | I | I | I | I | I | I | I |
| 10 | SO | I | I | I | I | I | I | I | I |
| 15 | SO | I | I | I | I | I | I | I | I |
| 20 | SO | O | G | G | G | G | G | G | G |
| 30 | SO | O | G | G | G | G | G | G | G |
| 40 | SO | SO | O | O | GG | GG | GG | GG | GG |
| 50 | SO | SO | O | O | GG | GG | GG | GG | GG |
| 55 | SO | SO | SO | O | O | GG | GG | GG | GG |
| 60 | SO | SO | SO | SO | O | O | GG | GG | GG |
| 70 | SO | SO | SO | SO | O | O | O | **GGG** | GG |
| 80 | SO | SO | SO | SO | O | O | O | GG | GG |
| 90 | SO | SO | SO | SO | SO | O | O | O | GG |
| 100 | SO | SO | SO | SO | SO | O | O | O | GG |
| 120 | SO | SO | SO | SO | SO | O | O | O | GG |
| 140 | SO | SO | SO | SO | SO | O | O | O | GG |
| 170 |  |  |  |  |  |  |  | O | GG |
| 200 |  |  |  |  |  |  |  | O | GG |

**Table 2. The design of experiments carried out to find an ideal (p, N). Initial and reconstructed silhouettes are visually compared to result in subjective assessments: I – inaccurate, O – oscillations, SO – strong oscillations, G – Good result, GG – very good result**

A last normalization operation is required for the genomes so that the phenotypes – silhouettes – are independent of a particular location, size or rotation and are only compared in terms of their shape, as discussed at the beginning of this paper.

Once the initial population has been created, the second phase of the IGA consists in the evolution of generations to create innovation.

As Kelly writes, "By using IGAs, we hope to allow designers to enhance their creativity through design space exploration" [11]. The individuals can reproduce among themselves and in this way, produce new solutions.

Our IGA is built upon four operators that influence the evolution of generations:

- *Selection*: decides which individuals will reproduce and create children.
- *Crossover*: builds a child's genome from two parent genomes.
- *Mutation*: changes a genome in a random way after the crossover.
- *Killing*: decides which individuals from the parents' population will survive in the new generation.

The selection and killing operators depend on the fitness values assigned by the user to each individual, i.e. to each car silhouette representing a design solution. The fitness value $f$ is a number between 0 and 6 according to the grade given by the user via an interface showing six individuals at the same time. The user browses through individuals from a generation and gives them a fitness value corresponding to his objective, where 0 is the worst and 6 is the best (see Figure 9).

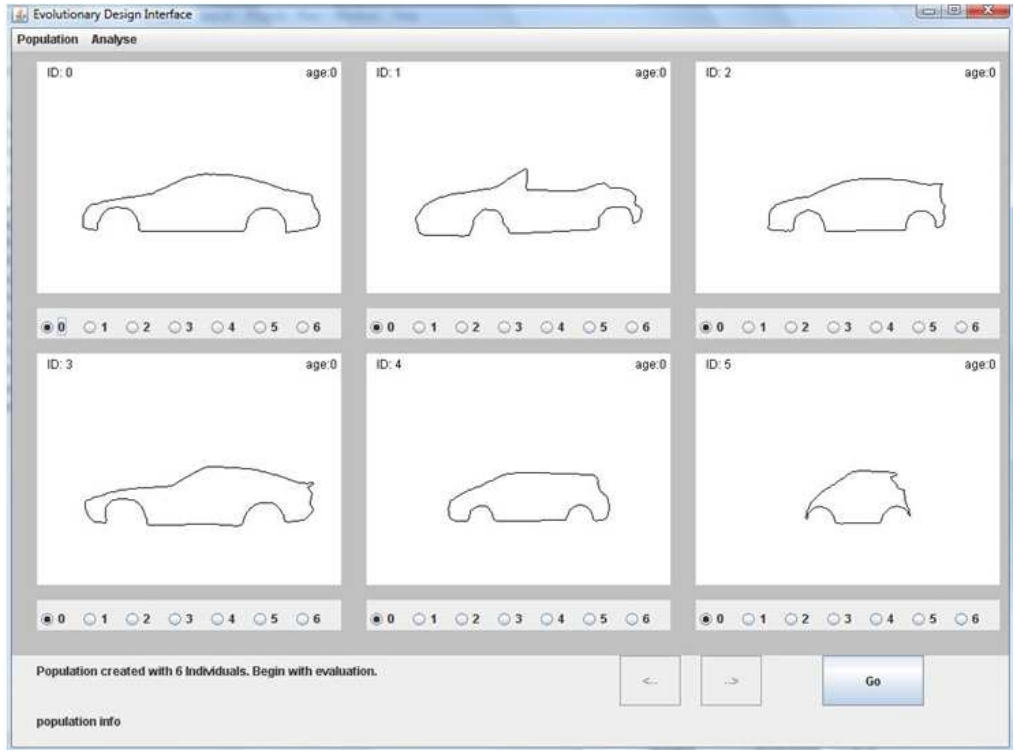

**Figure 9. The User Interface for the designer evaluation showing 6 individuals of a larger population. The designer can browse through the individuals by clicking on the arrow buttons**

We decided to adopt some conventional choices in term of selection and killing operators and to propose an original crossover operator. First, apart from the initial population of 30 individuals, we have fixed the number of individuals at 100 for each generation. We have chosen a turnover rate of 0.7, meaning that, for a coming generation, 30 individuals are kept from the previous one and 70 children are generated. In this way, potentially good design solutions are not lost. The probability for an individual to be selected as a parent is proportional to its fitness value (between 0 and 6). After two individuals are chosen from the parents' population, their genomes are combined into a child's genome by applying the crossover and mutation operators. Afterwards, the two individuals are put back into the parents' population. Indeed, the selection operator can select an individual more than once.

The crossover operator consists in a weighted average between the gene values of the two parents to build the genome of the child. A crossover weight W is chosen randomly between 0 and 100. A new gene $g^*$ is formed by calculating the weighted average of the genes $g_{m,1}$ and $g_{m,2}$ of the parents using formula (7).

$$g_m^* = \frac{W g_{m,1} + (100-W) g_{m,2}}{100} \tag{7}$$

In function of the weight *W*, we obtain different new design solutions which continuously interpolate a silhouette between the two parents' silhouettes (see Figure 11). In this way, a child greatly resembles its parents and produces almost no useless car solutions; for instance, the tires keep their rounded shapes. The disadvantage is the relatively small space explored for possible solutions: we stay in the convex of the initial individuals. Consequently, the population of design solutions tends to converge rapidly. To enlarge the space of possible solutions we must apply a mutation operator (not detailed here).

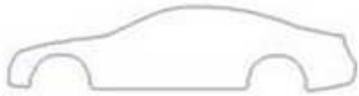

Figure 10. Two parent individuals

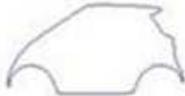

Figure 11. Results of a weighted average crossover between the genes of the two parents of Figure 10, using different weights W

Finally, the killing operator is applied to the original population and at first, kills all individuals who have a fitness of 0. These individuals are considered to be totally useless or totally unsatisfactory and shall no longer contribute to the evolution of the population. All other individuals have a chance of survival. The individuals to be killed are chosen by an inverse roulette wheel method. This means that the probability $pk_i$ for an individual to be killed can be expressed by formula (8).

$$pk_i = \frac{(7-f_i)}{\sum_{i=1}^{N^*}(7-f_i)} \tag{8}$$

where $f_i$ is the fitness of the individual $i$ and $N^*$ is the number of individuals in the population who have not been evaluated with a fitness of 0.

3. Similarity index : a validation tool

Is our system really capable of producing both innovation and novelty? With the help of our system, is it possible for a user to design a new car body silhouette which was not part of the initial population?

To answer these questions, it may be useful to have a tool that automatically measures the perceived difference between two car silhouettes and to prove whether the two car silhouettes are similar or not. We have then proposed to create a similarity index. Built upon user assessments, it measures the distance between two car profiles. The similarity index is an essential point in this study to perform tests and prove the relevance of the present evolutionary design framework to inspire designers. Two important criteria for the design of the index can be quoted:

- It must be mathematically founded.
- The calibration process must be stable and user-friendly.

The objectiveness issue is not a basic criterion: each designer can have his/her own conception of distance but it must, in all cases remain consistent from one assessment to another, when assessed by the same user (this is what we mean by the stability of the calibration process).

Hereafter, we propose a description of the process to obtain this similarity index. Two different models have been tested.

### 3.1. Mathematical definition

We first define $D(k,l)$, the distance between two genomes $G_k$ and $G_l$. Since modifications on the first ten genes only are significant (modifications on other genes do not change anything on the visual perception of the car profile), we hypothesize that the distance between two genomes $D(k,l)$ may be expressed as a weighted sum of the elementary distances between the first ten genes by formula (9).

$$D(k,l) = \sum_{m=1}^{10} \alpha(m) \parallel g_{k,m} - g_{l,m} \parallel^2 \qquad (9)$$

The factors $\alpha(m)$ are weighting factors that must give more importance to some genes in accordance with their participation in the apparent modification of the silhouettes. Here $g_{k,m}$ is gene number $m$ from genome $k$ and $g_{l,m}$ is gene number $m$ from genome $l$. One gene consists of two harmonics, called $a_m$ and $a_{-m}$, which are complex numbers. They can be written as: $a_m = u_m + i.v_m$.

Then, we define an elementary distance between two genes of the same order with formula (10).

$$\parallel g_{k,m} - g_{l,m} \parallel^2 = \frac{(u_{k,m}-u_{l,m})^2}{(u_{max,m}-u_{min,m})^2} + \frac{(u_{k,-m}-u_{l,-m})^2}{(u_{max,-m}-u_{min,-m})^2}$$

$$+ \frac{(v_{k,m}-v_{l,m})^2}{(v_{max,m}-v_{min,m})^2} + \frac{(v_{k,-m}-v_{l,-m})^2}{(v_{max,-m}-v_{min,-m})^2} \qquad (10)$$

where $u_{max,k}$ and $u_{min,k}$ (respectively $v_{max,k}$ and $v_{min,k}$) are the maximal and the minimal values of $u_k$ and $v_k$ on the whole initial population.

Finally, the similarity index is defined between two genomes $k$ and $l$ as expressed in formula (11).

$$SimInd\,(k,l) = \frac{100}{1+D(k,l)}\% = \frac{100}{1+\sum_{m=1}^{10}\alpha(m)\parallel g_{k,m}-g_{l,m}\parallel^2}\% \qquad (11)$$

So, with this definition, the similarity index is included between 0 and 100%, where 100% means that the two individuals are identical.

We now have to define the factor series $\alpha(m)$. Two solutions have been tried: an exponential form and a series of independent weights. These solutions are based on different assumptions and have been tested using a comparative experiment.

### 3.2. Exponential form

In this section we assume that $\alpha(m)$ can be written as an exponential expression which gives more importance to the first genes than to higher order genes, since a modification to the first genes impacts the car silhouette more than a modification to the last ones. So, $\alpha(m)$ is expressed by formula (12).

$$\alpha(m) = a\, e^{bm} \tag{12}$$

where *a* and *b* are two constant terms. So *D(k,l)* is now given by formula (13).

$$D(k,l) = \sum_{m=1}^{10} a\, e^{bm} \parallel g_{k,m} - g_{l,m} \parallel^2 = a \times \sum_{m=1}^{10} e^{bm} \parallel g_{k,m} - g_{l,m} \parallel^2 \tag{13}$$

But now we need to find the significant values of *a* and *b*.

We propose the following process to measure *b*:

- To choose a genome, which is copied 3 times: *G0*, *G1*, *G2*.
- To choose a gene *i* in *G1* (more significant with a low weight): $g1_i$.
- To choose a gene *j* (*i ≠ j*) in *G2* (more significant with a low weight): $g2_j$.
- To modify the gene $g1_i$ of *G1* in an arbitrary way.
- To modify the gene $g2_j$ of *G2* in such a way that there is an *iso-similarity* (defined below) between *G0* and *G1* on the one hand, and *G0* and *G2* on the other hand.

The *iso-similarity* is defined as follows: Two pairs of car silhouettes are *iso-similar* if the perceived level of similarity is the same for the two pairs. For example, here this means that the level of similarity is the same between *G0* and *G1*, and between *G0* and *G2*. Practically, it means that the user has to modify gene $g2_j$ until the level of perceived similarity becomes the same between *G0* and *G1* as between *G0* and *G2*. In this way, *G2* and *G1* are not identical, but their level of dissimilarity compared to *G0* is the same.

Then, this experiment results in equality as given in formula (14).

$$\alpha(i) \times \parallel g0_i - g1_i \parallel^2 = \alpha(j) \times \parallel g0_j - g2_j \parallel^2 \tag{14}$$

And *b* is finally given by formula (15).

$$b = \frac{1}{(j-i)} \times \ln \frac{\parallel g0_i - g1_i \parallel^2}{\parallel g0_j - g2_j \parallel^2} \tag{15}$$

By performing these tests *n* times with different car profiles, different users, and different *i* and *j* values, we get *n* different *b* values. The final value of *b* adopted is the average.

The next step consists in measuring *a*. We propose the following process: for each of the previous comparisons (between *G0*, *G1* and *G2*), the user defines the level of similarity $x\%$ ("the similarity between G0 and G1 on the one hand and G0 and G2 on the other hand is 70%" for example). To simplify the designer's evaluation task, the designer is asked to provide a similarity judgment on a 7-level scale which is converted into the percentage value as shown in Table 3.

The relation linking $x\%$ and *a* is given by formula (16).

$$x\% = \frac{100}{1 + a \times \sum_{m=1}^{70} e^{bm} \parallel g0_m - g1_m \parallel^2} \tag{16}$$

where *b* is the average value of the previous tests and *a* is finally obtained by formula (17).

$$a = \frac{1}{\sum_{m=1}^{70} e^{bm} \parallel g0_m - g1_m \parallel^2} \times \left(\frac{100}{x\%} - 1\right) \tag{17}$$

We also obtain 2*n* different values of *a*. The average value is acceptable if the standard deviation is low. The calculation of the similarity index is now completed.

| Level of similarity | Value of similarity index |
|:---:|:---:|
| 0 | 5% |
| 1 | 30% |
| 2 | 50% |
| 3 | 65% |
| 4 | 80% |
| 5 | 90% |
| 6 | 100% |

**Table 3. Scale of similarity for user assessments**

Practically, we use a Java interface (see Figure 12) that follows the processes described above through a series of random trials on individuals from the initial populations and with *i* and *j* values.

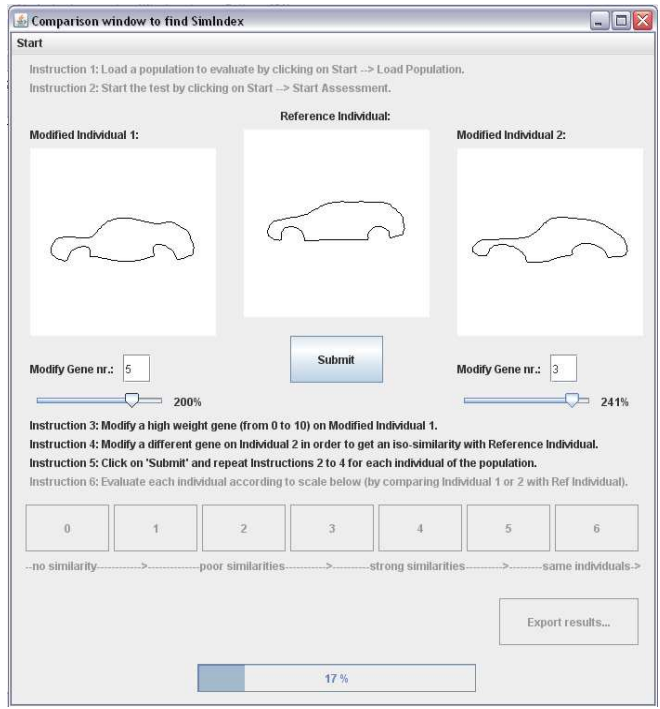

**Figure 12. Java interface to calculate parameters *a* and *b***

### 3.3. Weighted form

In the previous section, we assumed that $\alpha(m)$ should give more importance to the first genes than the last ones and can be written as an exponential expression. Let us ignore this assumption and try to associate each $\alpha(m)$ to a given weight $p_m$. Actually if this assumption easily seems true for the 3 or 4 first genes, it may be wrong for the next ones as it visually become harder to rank the influence of a modification on the (for instance) 5[th] or 6[th] harmonics.

So *D(k,l)* is expressed by formula (18).

$$D(k,l) = \sum_{m=1}^{10} p_m \parallel g_{k,m} - g_{l,m} \parallel^2 \tag{18}$$

To obtain the weights values, we follow the same process as with the exponential form. We finally obtain expressions linking weights $p_i$ and $p_j$ as in formula (19).

$$p_i = \frac{\|g0_j - g2_j\|^2}{\|g0_i - g1_i\|^2} \cdot p_j \quad (19)$$

By performing this test *n* times (*n>10*) with different values for *i* and *j* (to cover all of the first ten genes), we have a system of *n* equations, that can easily be resolved with the *logarithmic least square* method.

The tests are performed with a modified version of the Java interface as described in the previous section.

### 3.4. Choice of the index form

To find the best version of the similarity index, we performed the tests as described in Figure 13. 30. Cars were extracted from the initial population and were used to construct the two similarity indexes (calculation of *a* and *b* in the first case, calculation of the weights $p_i$ in the second case). 20 supplementary individuals from the initial population were then used to choose the most accurate index by calculating a similarity matrix. These matrices were compared to a third matrix containing the evaluation of the same 20 car silhouettes directly made by the users. These three matrices led to the calculation of the RMSE (Root Mean Square Error) associated with the two index versions.

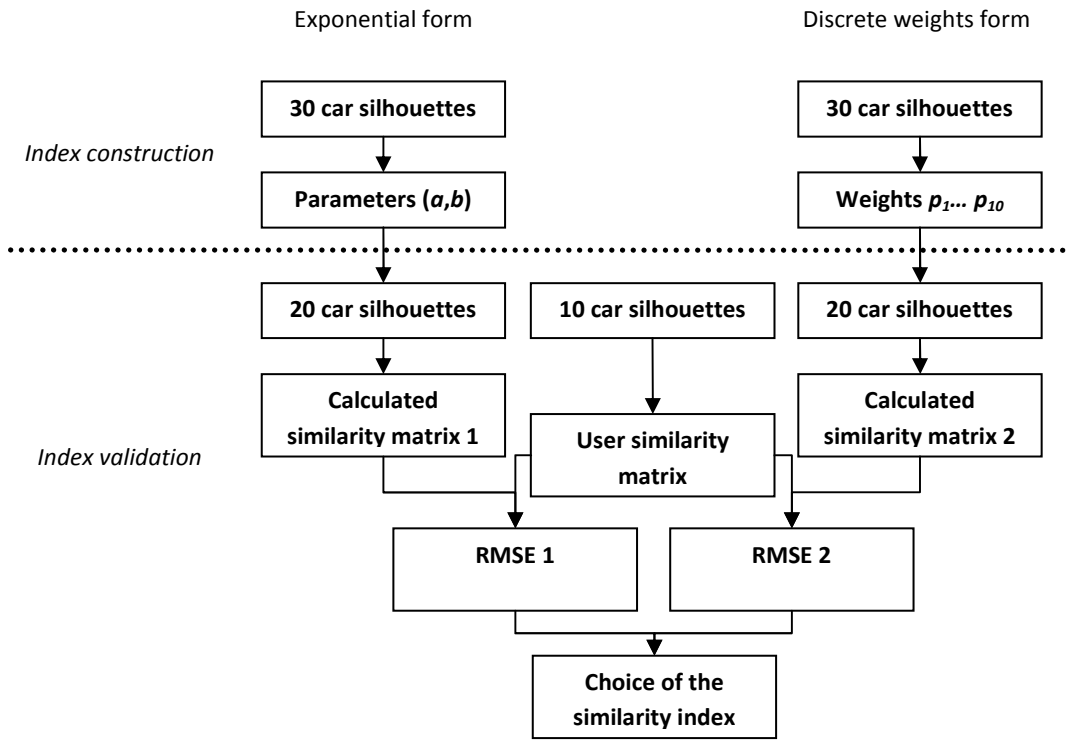

**Figure 13. Choice of the similarity index type**

We found an average RMSE of 7.04 for the exponential form index, and 31.82 with the independent weight form. Furthermore, in the first case we obtained a MAE (Mean Absolute Error) of 4.51, which shows a good precision for the model compared to the evaluation scale (0 to 100%). As such, the exponential form has been chosen for the similarity index. It has then been used to validate our model thanks to several user tests presented in the next section.

## 4. Model validation

Several tests have been performed to validate the model. We propose a short synthesis of some of these tests before detailing the ones that particularly prove that this evolutionary design schema stimulates creativity.

4.1 Synthesis of the validation tests

These first tests are presented in more detail in [24] and [25]. In this section, we only draw the main conclusions and findings concerning user satisfaction and the convergence of the model. The next section details three tests that mainly focus on creativity.

4.1.1 User satisfaction

The first way to validate the model consists in observing the evolution of user satisfaction along the generations. A test inspired by [17] has been performed by asking 8 users to evaluate 10 successive generations. These users are students in the design engineering field in the last year of master class. 4 users were asked to find their best *sportive* car profile, while the 4 other users worked with the *friendly* semantic attribute (find their best *friendly* car silhouette). We asked the users to perform hedonistic assessments (i.e. including their own preference). The mutation probability and the selection rate were respectively set to 0.05 and 0.7.

The average fitness values evolution on the 10 generations (see Figure 14) proves that the users are satisfied by the model. Indeed, the average user fitness increases from 2.3 to 3.9 for the semantic attribute sportive, and from 3.0 to 5.0 for the *friendly* attribute, i.e. an increase of about 70% in the two cases.

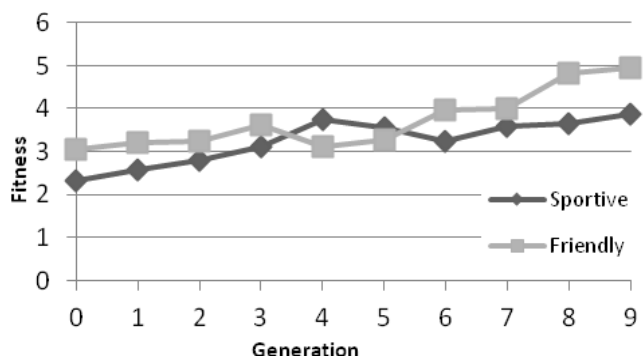

Figure 14. Fitness evolution for the convergence test

4.1.2 Convergence

To assess the ability of the model to converge, two tests based on the data obtained with the previous 8 users and the similarity index have been proposed. They aim to study the evolution along the first ten generations of the average similarity between individuals for each user (for the first test) and between the users (for the second one). In addition, we study the similarities that exist between the best individuals from the last generation for each designer.

On average, a similarity of 6.5% has been calculated for the initial population. For all users, this average value increases rather quickly with the generations, reaching for the entire user range a similarity between 60 and 90% at the $10^{th}$ generation. This qualitatively means that it is hard to visually perceive strong differences between individuals from the last generation. It shows that the IGA is able to converge towards a single individual family.

Moreover, the study of similarity between the best individuals from the last generation for all users shows important results:

- For the "sportive" semantic attribute, the similarity between the best individuals of the last generation gives a similarity of about 51%: all users converge towards the same individual family (see Figure 15). However, it is not possible to identify if the behavior is due to the choice of users or the model that does not converge towards another family of profiles. It is also useful to note that this common family is very close to a Porsche 911 profile, which shows that this profile is probably perceived by the users as the ideal sportive car silhouette.

- For the "friendly" semantic attribute, the similarity between the best individuals from the last generation is only 9.21%: there are only a few connections between the individual families reached by the different users (see Figure 16). It answers a previous question: the model preserves the freedom of thinking for each designer, who is able to converge towards his/her own mental target.

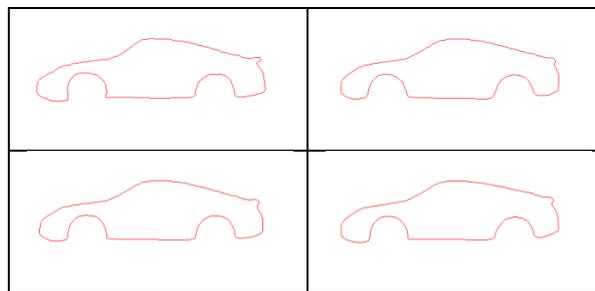

Figure 15. Examples of best "sportive" car silhouettes (one from each user)

The results of these first tests show that the IGA has a really promising behavior. However, some crucial questions remain and further tests are required to complete a full validation. The main questions are as follows:

- Is the model really able to create novelty, or are the final car profiles that emerge always correlated to some individuals from the initial population?

- How can we be sure that the IGA obtains better results compared to a more classical (manual) method? In other words, what is the real added value of the IGA?

In the next sections, new tests are proposed that mainly focus on novelty emergence and creativity enhancing, which are essential aspects for innovation.

4.2 Novelty emergence

A major element to enhance creativity is the model's ability to create novelty, i.e create any individual, randomly chosen or imagined by the user. In other words, the purpose of this test is to show that it is possible to reach a defined individual which is not part of the initial population.

To answer this question, we can execute a simple test. A subject draws a car body silhouette on a sheet of paper which comes spontaneously to his or her mind and which is not part of the initial population. This car body silhouette is taken as a "target individual". Using our proposed IGA, the user must now try to obtain the "target silhouette" he has drawn on a sheet of paper by the end of the IGA process. Of course, the objective of this test is to prove that there is no limit to our system to converge towards the imagined shape. Then, keeping this reference drawing in front of him or her, the user evaluates the car contour individuals, providing higher grades to contours which are similar to the target contour and lower grades to those that are dissimilar. By considering both the number of generations required to resemble the target individual and the similarity value, we can estimate the quality of our design system.

Alternatively, the target car silhouette may be an individual from the initial population which is removed from this initial population.

We have preferred to make abstraction of the designer subjectivity by automating the ability of the system to converge towards an ideal car silhouette, in order to measure the sole quality of the method. The role of the designer is played by a similarity fitness calculation algorithm, which automatically evaluates individuals from a generation in terms of their similarity to the target individual, thanks to the similarity index previously defined.

This test has been repeated several times, with different target individuals extracted from the initial individual population for an average final similarity index of 90%. Let us provide a practical example with the target individual given in Figure 17.a. The genetic algorithm parameters were as follows: population of 100 individuals, turnover rate of 0.7 and mutation probability of 0.3. The mutation can change a gene within a range of ±(50%-200%). After 10 generations, our system reached the car body silhouette in Figure 17.b which has a similarity index of 92%, and which can be visually considered as a much more satisfactory result.

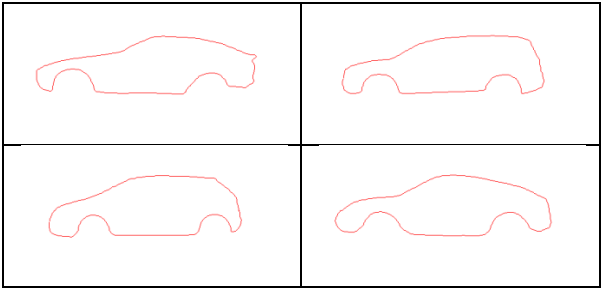

Figure 16. Examples of best "friendly" car silhouettes (one from each user)

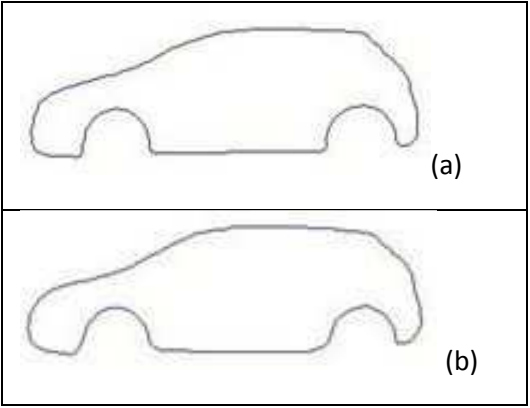

Figure 17. Comparison between the reference silhouette (a) and the final resulting silhouette (b)

The average fitness of the population converges over the generations to a high value (see Figure 18), whereas the value of the best similarity index in the population (the fitness of the fittest individual) rises rapidly from a relatively low 44% to 92%.

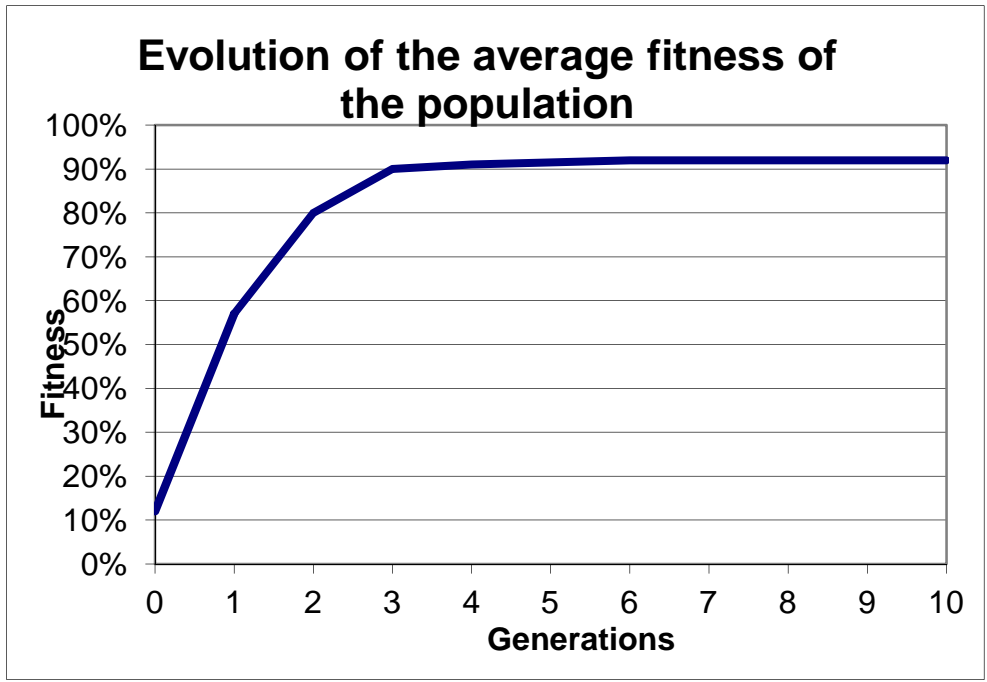

**Figure 18. The average fitness of all individuals in the population over the generations**

|  | **Best paper individuals** | | |
|---|---|---|---|
| | 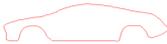 | 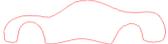 | 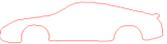 |
| **Best individuals from the IGA** 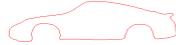 | >> | >>> | > |
| 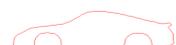 | > | >>> | = |
| 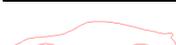 | > | >>> | = |

**Figure 19. Example of pairwise comparison matrix used for the tests**

Furthermore, another test has been performed to study the link between the best individuals from the 10[th] generation identified by the users and the initial population. The results are synthesized in Table 4 for User 2 working with the "sportive" semantic attribute. It shows that his/her best car profiles are highly inspired by the 13[th] individual from the population, namely the Porsche 911 silhouette. However, no correlation is observed with the "friendly" semantic attribute, and it can be concluded that the correlation between the Porsche 911 and the results of the users with the "sportive" semantic attribute are the consequence of a choice (voluntary or not), and not a constraint of the model.

|  |  | User 2 | | |
| --- | --- | --- | --- | --- |
|  |  | Best Individual 1 | Best Individual 2 | Best Individual 3 |
| Initial Population | Ind 0 | 2,04 | 1,93 | 1,84 |
| | Ind 1 | 2,95 | 2,77 | 2,65 |
| | Ind 2 | 10,02 | 9,02 | 8,80 |
| | Ind 3 | 0,91 | 0,88 | 0,90 |
| | Ind 4 | 3,45 | 3,24 | 3,35 |
| | Ind 5 | 2,04 | 1,94 | 1,91 |
| | Ind 6 | 1,42 | 1,36 | 1,36 |
| | Ind 7 | 1,78 | 1,74 | 1,52 |
| | Ind 8 | 13,90 | 13,73 | 10,75 |
| | Ind 9 | 6,62 | 6,17 | 6,53 |
| | Ind 10 | 2,76 | 2,60 | 2,47 |
| | Ind 11 | 3,28 | 3,41 | 2,91 |
| | Ind 12 | 2,08 | 1,97 | 1,97 |
| | Ind 13 | 69,65 | 69,89 | 53,21 |
| | Ind 14 | 3,50 | 3,30 | 3,46 |
| | Ind 15 | 7,15 | 7,23 | 10,83 |
| | Ind 16 | 2,07 | 1,97 | 1,93 |
| | Ind 17 | 3,05 | 2,88 | 2,87 |
| | Ind 18 | 1,65 | 1,58 | 1,47 |
| | Ind 19 | 1,99 | 1,93 | 1,69 |

Table 4. Similarity values between the best individuals and the initial population for the "sportive" semantic attribute (in %)

Finally, the results of these two tests show that the final individual is genetically very similar to the target individual. The IGA is then able to create totally new individuals that are not part of the initial population. For us, this is a necessary condition to improve creativity. However, further tests are required to show that our model brings added value compared to a classical "manual" method.

4.3 Creativity enhancing

To show that the results obtained with the IGA are better than without the IGA, we have performed two different tests from a same base inspired by Kim and Cho's works [17].

The same set of data as used for some of the previous tests was used (8 users during 10 generations, with 2 semantic attributes). However, we also asked the user to evaluate 400 randomly generated individuals with the same fitness scale (between 0 and 6). These individuals were printed on a large sheet of paper and numbered. An Excel routine was given to the users to identify, in several evaluation steps, their three preferred profiles according to their assigned semantic attribute ("friendly" or "sportive"). To clarify the explanations in the rest of the document, these best individuals are called *paper individuals*.

The users were then asked to compare the three best paper individuals with the three best individuals from the 10[th] generation of the IGA, thanks to a pairwise comparison matrix. The comparisons were performed according to the scale presented in Table 5. As such, users filled the matrix with the mathematical symbols (>, >>, =...) that were then replaced by numbers to analyze the data.

| -3 | <<< | highly inferior |
|----|-----|-----------------|
| -2 | <<  | inferior        |
| -1 | <   | slightly inferior |
| 0  | =   | equal           |
| 1  | >   | slightly superior |
| 2  | >>  | superior        |
| 3  | >>> | highly superior |

**Table 5. Hedonistic (preference) scale of pairwise comparisons of car profiles**

Considering this data, the first test evaluates the user satisfaction of results obtained with the IGA. The second test analyses the data with the similarity index.

Table 6 shows the test results for each user and for each semantic attribute.

| | | |
|---|---|---|
| ***Sportive*** | User 1 | +1,78 |
| | User 2 | +1,67 |
| | User 3 | +1,56 |
| | User 4 | +1,00 |
| | Average | +1,50 |
| ***Friendly*** | User 5 | -0,33 |
| | User 6 | +0,67 |
| | User 7 | +0,78 |
| | User 8 | +1,22 |
| | Average | +0,58 |
| **Total Average** | | **+1,04** |

**Table 6. Results of the pairwise comparisons between best IGA solutions and best paper (random) solutions highlighting the superiority of IGA**

Some details have to be explained to understand the meaning of Table 6. Pairwise comparisons are made in accordance with the following scheme: the best individuals from the IGA are compared to the best paper individuals. For example, User 4 has an average evaluation of +1.00. According to Table 5, this means that User 4 found the best individuals from the IGA slightly superior to the best paper individuals. Thus, positive numbers prove that the results obtained with our model are better than without.

The results show good behavior for the *sportive* semantic attribute: the average score is +1.50, so the best IGA individuals are between "slightly superior" and "superior" to the best paper individuals.

For the *friendly* attribute, the results show good behavior as well, even if the difference is slightly less perceptible. All users prefer individuals from the IGA except user 5, who prefers the individuals without the model. Globally, the results for this attribute are not really homogenous, and can be explained by a more subjective comprehension (and as such, characteristic to each user) of the word *friendly* rather than the word *sportive* (this observation is highlighted in other tests presented hereafter).

Now, let us compare if these two sets of 3 best solutions can have close solutions or not. The maxima of similarity are shown in Table 7. For 5 users out of 8, the maximum similarity is above 70%. For 2 users, the maximum is between 20 and 30%. For the last user, the similarity is only 2.67% .

|  | | |
|---|---|---|
| ***Sportive*** | User 1 | 84,49 |
|  | User 2 | 21,26 |
|  | User 3 | 80,92 |
|  | User 4 | 97,93 |
|  | Average | 71,15 |
| ***Friendly*** | User 5 | 2,67 |
|  | User 6 | 98,09 |
|  | User 7 | 71,32 |
|  | User 8 | 26,94 |
|  | Average | 49,75 |
| **Total Average** |  | **60,45** |

**Table 7. Maxima of similarity values between best individuals of the IGA and of paper individuals (in %)**

This means that, most of the time, the final solutions obtained are quite similar, whatever the design method (IGA system or brute force enumeration). Indeed with the IGA, without any judgment of value, 5 users have found *at least* one individual that is very close to one of the paper individuals.

But, 400 individuals were proposed on paper and it takes about 45 minutes for one user to evaluate them. Comparatively, less than the 200 individuals used with the IGA process (10 generations of 20 individuals, minus those who survive from one generation to the next), and the evaluation only lasts about 20 minutes. So, it appears that our system brings at least comparable results more quickly and with fewer individuals; this is clearly a proof of its utility.

## 5. Concluding remarks

Shah et al. [26] propose four separate measures to assess the effectiveness of a generation process towards innovation:

- Novelty, which is a measure of how unusual or unexpected a concept is, as compared to other concepts,

- Variety measures the size of the solution space,

- Quality explores the feasibility of a concept and how close it comes to meet the design specifications,

- Quantity concerns the total number of concepts generated.

Transposed to this study, these measures allow the creativity enhancing of the proposed IGA model to be highlighted.

We have shown that the model is able to create individuals that are not closely linked with individuals from the initial population. The model reaches a high level of novelty potential.

Concerning variety, the size of the solution space is theoretically unlimited since continuous genes are used.

Here, the quality measure can be related to the user satisfaction of the end-result of the design activity. Here, the superiority of designs generated by the IGA has been proved compared to a brute systematic exploration of shapes.

Finally, the quantity can be totally customized by the user: he/she chooses the size of the population, as well as the sensibility of the algorithm (killing and mutation factors). The choice of precise semantic attributes is a guide to generate different car profile concepts.

In addition, our model is extensible as defined by Gruber [21], since a designer may introduce, at any moment of the evolutionary algorithm, a customized profile he or she has just sketched, or that he/she locally modifies from an existing individual of the current generation. This facility has not been detailed here but completes the numerous advantages to further study Interactive Genetic Algorithms to explore spaces of 2D innovative curves.

## 6. Acknowledgments

We gratefully thank François Bleibel, Nicolas Cordier, Gilles Foinet and Thomas Ricatte for their contributions to this work.